\documentclass{article}
\usepackage{ijcai25}

\usepackage{times}
\usepackage{soul}
\usepackage{url}
\usepackage[hidelinks]{hyperref}
\usepackage[utf8]{inputenc}
\usepackage[small]{caption}
\usepackage{graphicx}
\usepackage{amsmath}
\usepackage{amsthm}
\usepackage{booktabs}
\usepackage{algorithm}
\usepackage{algorithmic}
\usepackage[switch]{lineno}
\usepackage{multirow}
\usepackage{makecell}
\usepackage{xcolor}
\usepackage{amssymb}
\usepackage{orcidlink} 

\title{Multi-modal Anchor Gated Transformer with Knowledge Distillation for Emotion Recognition in Conversation}

\author{
Jie Li$^{1,2}$\and
Shifei Ding$^{1,2}$\footnote{Corresponding Author.}\and
Lili Guo$^{1,2}$\And
Xuan Li$^{1,2}$\\
\affiliations
$^1$School of Computer Science and Technology, China University of Mining and Technology\\
$^2$Mine Digitization Engineering Research Center of Ministry of Education, China University of Mining and Technology\\
\emails
\{jie\_li, dingsf, liliguo, lixuan23\}@cumt.edu.cn,
}

\begin{document}

\maketitle

\begin{abstract}
Emotion Recognition in Conversation (ERC) aims to detect the emotions of individual utterances within a conversation. Generating efficient and modality-specific representations for each utterance remains a significant challenge. Previous studies have proposed various models to integrate features extracted using different modality-specific encoders. However, they neglect the varying contributions of modalities to this task and introduce high complexity by aligning modalities at the frame level. To address these challenges, we propose the Multi-modal Anchor Gated Transformer with Knowledge Distillation (MAGTKD) for the ERC task. Specifically, prompt learning is employed to enhance textual modality representations, while knowledge distillation is utilized to strengthen representations of weaker modalities. Furthermore, we introduce a multi-modal anchor gated transformer to effectively integrate utterance-level representations across modalities. Extensive experiments on the IEMOCAP and MELD datasets demonstrate the effectiveness of knowledge distillation in enhancing modality representations and achieve state-of-the-art performance in emotion recognition. Our code is available at: \url{https://github.com/JieLi-dd/MAGTKD}.
\end{abstract}

\section{Introduction}
\label{sec:intro}

\begin{figure} 
    \centering 
    \includegraphics[width=1\linewidth]{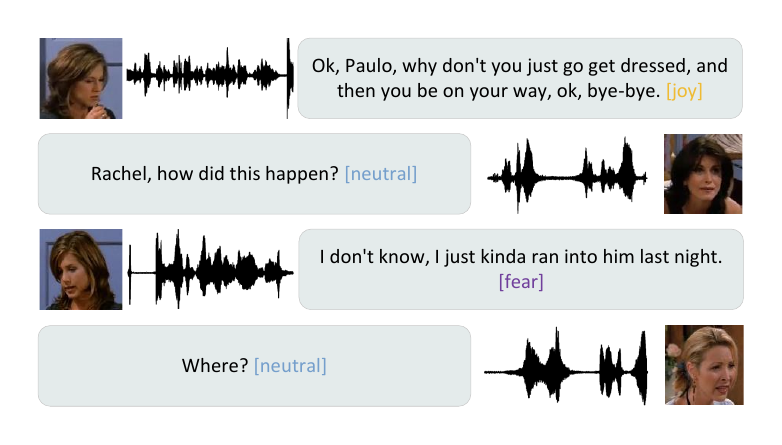} 
    \caption{Multi-modal conversation example from the MELD dataset.} 
    \label{MERC example} 
\end{figure}

Emotion plays a pivotal role in human communication, influencing not only the content but also the tone and context of interactions. Emotion Recognition in Conversation (ERC) aims to identify the emotional states expressed in each utterance within a dialogue. This task is essential for applications in areas such as conversational agents, healthcare systems, and recommendation engines. While emotions are traditionally expressed through text, they are also richly conveyed in the audio and visual modalities \cite{poria-etal-2017-context,Wu2025DEVA}. Figure \ref{MERC example} provides an illustrative example of multi-modal ERC, showcasing how information from various modalities can be integrated to improve emotion recognition.

Recent research in ERC has primarily focused on individual modalities. Text-based models often leverage context modeling \cite{Poria_2019_dialoguernn,song-etal-2022-supervised,hu-etal-2023-sacl,yang-2024-VAD-VAE} or incorporate external knowledge \cite{zhong-etal-2019-knowledge,zhu-etal-2021-topic,lee-lee-2022-compm,Wang2025KSDGCN}. Audio-based models make use of multi-task learning, attention mechanisms, and data augmentation strategies \cite{latif-2023-MTL-AUG,He2025Str-GCL,GUO2025APIN}, while video-based models extract key frames to enhance emotion recognition \cite{wei-2021-video-key,poria-etal-2017-context}. However, relying solely on a single modality for emotion recognition can overlook crucial emotional cues embedded in other modalities, leading to suboptimal performance. This limitation has spurred increasing interest in multi-modal ERC. Existing multi-modal models typically extract frame-level features for each modality, align these features across modalities, and fuse them for emotion classification \cite{tsai-etal-2019-multimodal,guo-2022-IA-MMTF,zheng-etal-2023-facial}. While effective, these approaches often treat all modalities equally, disregarding the varying significance of each modality in emotion recognition. Furthermore, the complex alignment process increases the computational burden, making these models less suitable for deployment in resource-constrained environments. To address these challenges, a more efficient and adaptive approach to modality representation and integration is needed.

Prompt-based learning has recently gained attention for its success in both natural language processing (NLP) and multi-modal tasks. In ERC tasks focusing on textual data, well-designed prompts can effectively guide models to extract relevant contextual information, thereby improving the quality of utterance-level features \cite{song-etal-2022-supervised,son-etal-2022-grasp,yun-etal-2024-telme}. Additionally, knowledge distillation techniques have been widely explored to enhance the performance of student models by enabling them to mimic better the representations learned by teacher models \cite{lin2022knowledge,li2023curriculum,yun-etal-2024-telme,Ma-2024-SDT}. To overcome the challenges in multi-modal ERC, we introduce the Multi-modal Anchor Gated Transformer with Knowledge Distillation (MAGTKD), a framework designed to improve the integration of multi-modal information for emotion recognition tasks. Specifically, MAGTKD leverages context-aware prompts to extract high-quality utterance-level textual representations. These robust textual features are then used within a knowledge distillation framework to enhance the representation capacity of weaker modalities (e.g., audio and video), ultimately improving the fusion of multi-modal features for emotion recognition. In contrast to existing methods, which rely on frame-level feature interactions before fusion—thereby increasing computational complexity \cite{tsai-etal-2019-multimodal,guo-2022-IA-MMTF,zheng-etal-2023-facial}—MAGTKD directly fuses utterance-level features post-interaction, addressing the computational burden while maintaining or even improving performance.

We evaluate MAGTKD on two widely-used benchmark datasets, IEMOCAP and MELD. Experimental results demonstrate that MAGTKD achieves state-of-the-art performance on both datasets, surpassing existing methods in both accuracy and efficiency.

The key contributions of this work are as follows: 
    \begin{itemize} 
        \item We propose MAGTKD, a novel framework for ERC that effectively integrates multi-modal features, taking into account the varying contributions of different modalities to emotion classification. 
        \item MAGTKD significantly reduces model complexity compared to traditional frame-level feature fusion methods. 
        \item MAGTKD sets new benchmarks for ERC, achieving superior performance on the IEMOCAP and MELD datasets. 
    \end{itemize}

\section{Related Works}
\label{sec: related works}

\subsection{Prompt Learning}
Prompt Learning has emerged as an effective approach for leveraging pre-trained models by designing task-specific prompts to fine-tune and integrate them for downstream tasks, enabling improved modality representations. It has been widely adopted across various NLP tasks \cite{gao-etal-2021-making,heinzerling-inui-2021-language,xu-etal-2023-efficient}. Recently, researchers have begun exploring the application of prompt learning in multi-modal settings \cite{Tsi-2021-multimodal-few-shot,khattak2023maple,zhu2023visual}. \cite{Tsi-2021-multimodal-few-shot} presents a simple, yet effective, approach for transferring this few-shot learning ability to a multi-modal setting (vision and language). \cite{khattak2023maple} proposes Multi-modal Prompt Learning (MaPLe) for both vision and language branches to improve alignment between the vision and language representations. \cite{zhu2023visual} develop Visual Prompt multi-modal Tracking (ViPT), which learns the modal-relevant prompts to adapt the frozen pre-trained foundation model to various downstream multi-modal tracking tasks. With success in diverse NLP and multi-modal learning applications, we extend prompt learning to the emotion recognition task, aiming to harness its potential for enhancing emotional feature extraction and representation.

\subsection{Konwledge Distillation}
Knowledge Distillation (KD) aims to transfer knowledge from a large teacher network to a smaller student network. This knowledge transfer typically occurs at three levels: soft labels of the final layer \cite{KD:journals/corr/HintonVD15}, intermediate-layer features \cite{DBLP:iclr/corr/RomeroBKCGB14}, and the relationships between features across layers \cite{yim2017gift}. Based on the learning strategy, KD can be categorized into offline \cite{passalis2018learning,li2020few} and online \cite{zhang2018deep,chung2020feature} distillation. In offline distillation, the teacher model is pre-trained to guide the student model's learning. In contrast, online distillation involves simultaneous training of the teacher and student models with joint parameter updates. KD has demonstrated its effectiveness in transferring knowledge across modalities in multi-modal research \cite{albanie2018emotion}. Motivated by this, we adapt KD techniques to the multi-modal ERC task, enabling efficient knowledge transfer between modalities to enhance emotion recognition performance.

\subsection{Modal Fusion}
In the domain of modality fusion, existing works predominantly focus on extracting frame-level features and performing feature interactions at this granularity. \cite{tsai-etal-2019-multimodal} introduces the Multi-modal Transformer (MulT) to generically address the above issues in an end-to-end manner without explicitly aligning the data. \cite{zheng-etal-2023-facial} extracts three modal frame-level features and uses an attention mechanism to perform alignment operations on the three modal features. However, frame-level feature alignment often introduces significant computational complexity. Unlike these approaches, our work adopts utterance-level feature extraction and designs a novel model for multi-modal feature fusion, effectively reducing complexity while maintaining strong performance in emotion recognition tasks.

\section{Methods}
\begin{figure*}
    \centering
    \includegraphics[width=1\linewidth]{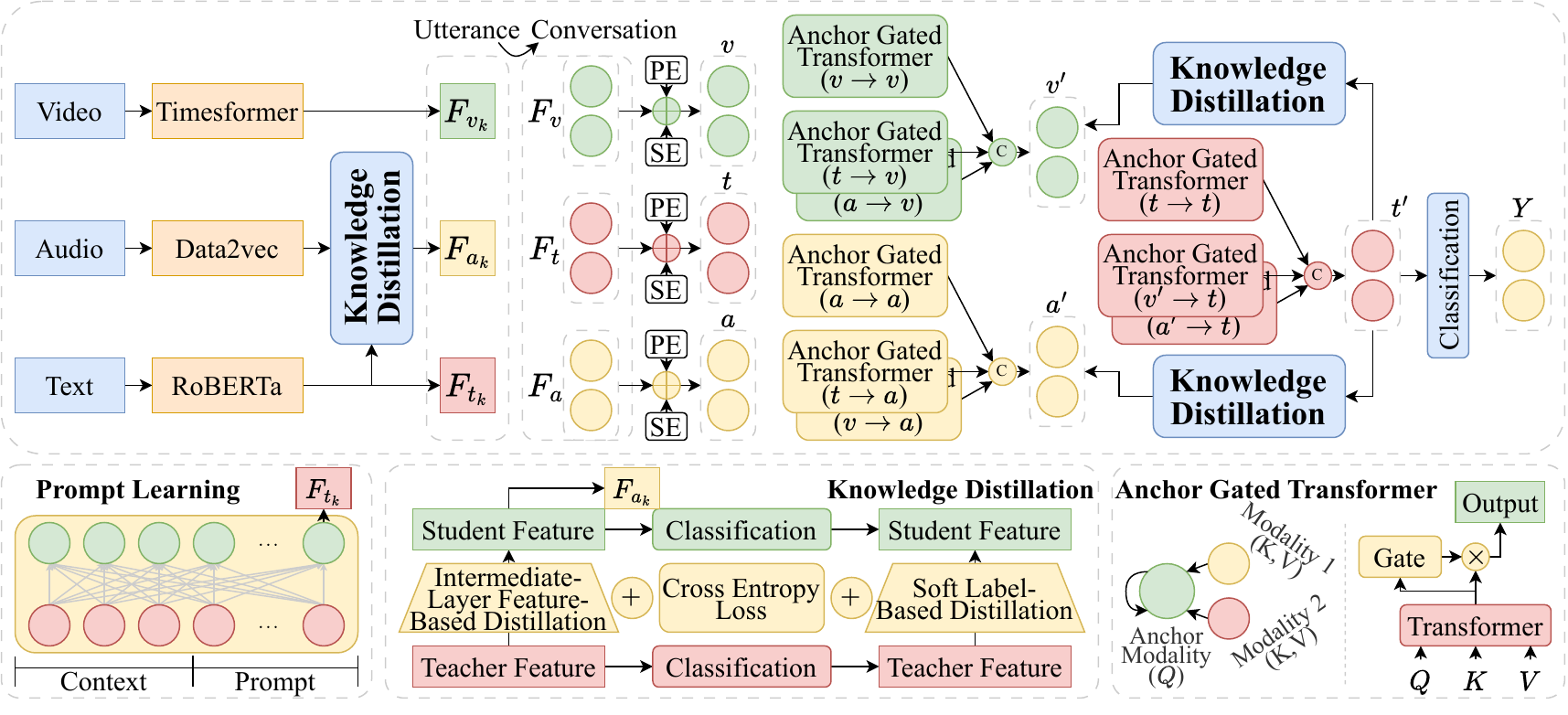}
    \caption{Illustration of the architecture of MAGTKD.}
    \label{framework}
\end{figure*}

To enhance the representation of each modality and achieve effective multi-modal fusion, we propose the MAGTKD model for the ERC task. Figure \ref{framework} illustrates the overall architecture of the proposed framework, with detailed descriptions provided in the following subsections. 

\subsection{Task Definition}
Given a set of speakers $S$, utterances $U$, and emotion labels $Y$, a conversation consisting of $k$ utterances is represented as $[s_i,u_1,y_m,s_j,u_2,y_n,...,s_i,u_k,y_m]$, where $s_i,s_j \in S$ are speakers and $y_m,y_n \in Y$ are one of the predefined emotion categories. If $i=j$, $s_i$ and $s_j$ represent the same speaker. Moreover, $u_k \in U$ represents the $k$-th utterance. Each utterance $u_k=\{t_k,a_k,v_k\}$ contains three modalities, where $t,a,v$ represent text, audio and video, respectively. The goal of ERC is to predict to which emotion label $y_m$ the utterance $u_k$ belongs.

\subsection{Feature Extraction}
Figure \ref{framework} illustrates the extraction of modality-specific features using dedicated encoders for each input modality. In this section, we explain the feature generation process in detail:

\textbf{Text}: Following prior work \cite{lee-lee-2022-compm,song-etal-2022-supervised}, we utilize prompt learning to model both the context and speaker information. For the text encoder, we adopt RoBERTa \cite{liu-2019-roberta}. We construct the contextual representation $C_k$ and the prompt $P_k$ as follows.
\begin{equation}
    C_k=Concat(s_i: t_1, s_j: t_2, \dots, s_i: t_k)
\end{equation}
\begin{equation}
    P_k=For ~ s_i: t_k ~ Now ~ s_i ~ feels ~ <mask>
\end{equation}
\begin{equation}
    F_{t_k} = Roberta(C_k~</s>~P_k)
\end{equation}
where $<mask>$ represents a special token, $F_{t_k} \in R^{1 \times d}$ is embedding of $<mask>$, representing the aggregated emotion feature, and $d$ is the hidden dimension of the $<mask>$ token.

\textbf{Audio}: Self-supervised learning has achieved remarkable success not only in natural language processing but also in audio and video domains \cite{Bae-2020-wav2vec,baevski2022data2vec}. For the audio encoder, we employ Date2vec \cite{baevski2022data2vec}, with the audio segment $a_k$ of the $k$-th utterance as input. The process of extracting audio features is formalized as follows.
\begin{equation}
    F_{a_k}=Data2vec(a_k)
\end{equation}
where $F_{a_k} \in R^{1 \times d}$ is the embedding of $a_k$, and $d$ is the hidden dimension of audio features.

\textbf{Video}: Similar to the audio feature extraction process, we utilize Timesformer \cite{bertasius2021space} as the video encoder. The process of extracting video features is formalized as follows.
\begin{equation}
    F_{v_k}=Timesformer(v_k)
\end{equation}
where $v_k$ is the video input, and $F_{v_k} \in R^{1 \times d}$ is the embedding of $v_k$, and $d$ is the hidden dimension of video features.

\subsection{Knowledge Distillation}
Unlike traditional knowledge distillation methods that utilize KL divergence, the multi-modal ERC task involves cross-modal knowledge distillation. We adopt a collaborative distillation strategy based on soft labels and intermediate-layer features, using Pearson correlation coefficients as our cross-modal measurement approach.
\begin{equation}
    d(u,v)=1-p(u,v)
\end{equation}
where $p(u,v)$ is the Pearson correlation coefficient between two logit vectors $u$ and $v$.

Soft Label-Based Distillation leverages the soft label outputs from the last layer of each modality encoder to compute the knowledge divergence across modalities at both the sample and feature levels. Pearson correlation coefficients are employed to measure the degree of knowledge disparity between different modalities. By reducing this disparity, the textual features transfer knowledge to the audio features. The process is formalized as:
\begin{equation}
    Y_{i,:}^{t}=softmax(P_{i,:}^{t}/\tau)
\end{equation}
\begin{equation}
    Y_{i,:}^{a}=softmax(P_{i,:}^{a}/\tau)
\end{equation}
\begin{equation}
    L_s = \frac{\tau^2}{B}\sum_{i=1}^{B}d(Y_{i,:}^{a},Y_{i,:}^{t}) + \frac{\tau^2}{C}\sum_{j=1}^{C}d(Y_{:,j}^{a},Y_{:,j}^{t}) \label{loss_s}
\end{equation}
where $B$ is a training batch, $C$ is the emotion categories, $P^t, P^a \in R^{B \times C}$ are the prediction matrix of text and audio modality, respectively. $\tau$ is a temperature parameter to control the softness of logits.

Intermediate-Layer Feature-Based Distillation computes similarity matrices within a batch for the textual modality as the target matrix (via dot product between text modality features and their transpose). Similarly, a source matrix is computed for the audio and text modalities. Using the softmax function, we derive the target and source distributions. The KL divergence between these distributions is minimized to enable the transfer of knowledge from textual features to audio features. The process is defined as:
\begin{equation}
    T_i = \frac{exp(F_{i,j}/\tau)}{\sum_{s=1}^{B}exp(F_{i,s})},\forall{i,j \in B}
\end{equation}
\begin{equation}
    S_i = \frac{exp(F_{i,j}^{'}/\tau)}{\sum_{s=1}^{B}exp(F_{i,s}^{'})},\forall{i,j \in B}
\end{equation}
\begin{equation}
    L_f = \frac{1}{B}\sum_{i=1}^{B}KL(T_i || S_i)  \label{loss_f}
\end{equation}
where $F_{i,j},F_{i,j}^{'} \in R^{B \times B}$ are the text-modal similarity matrix and the text-audio modal similarity matrix, respectively. $T_i,S_i$ are target and source distributions.

The overall loss function includes the above two losses and the cross-entropy loss:
\begin{equation}
    L_{CE} = -\frac{1}{B}\sum_{i=1}^{B}y_{i}\cdot \log p_i
\end{equation}
\begin{equation}
    L_{all} = L_{CE} + L_s + L_f \label{Loss1_all}
\end{equation}
where $y_i$ is true labels and $p_i$ is predict labels.

\subsection{Multi-modal Anchor Gated Transformer}
\begin{figure}
    \centering
    \includegraphics[width=1\linewidth]{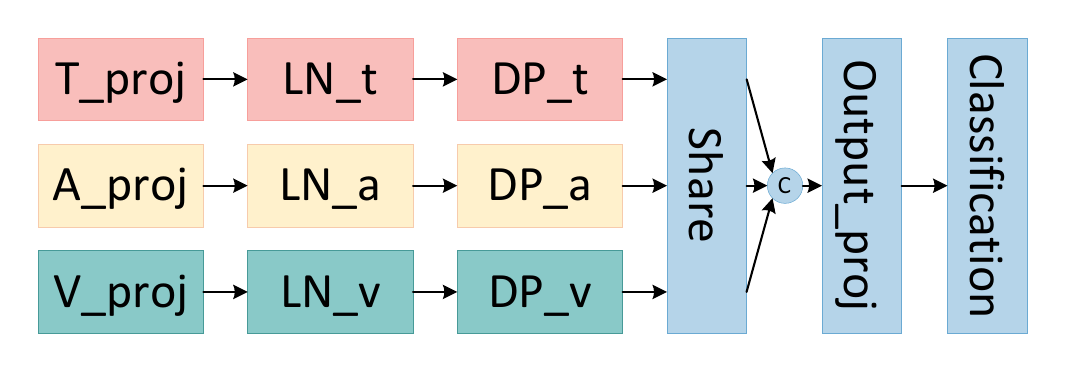}
    \caption{Architecture of the Concat model. Each modality passes through a linear layer, followed by Layer Normalization (LN) and Dropout (DP). The outputs are then processed by a shared weights module before classification through a final linear layer.}
    \label{fig:Concat}
\end{figure}

In the first stage, we utilize prompt learning and knowledge distillation to extract utterance-level features for each modality. However, directly concatenating these features for emotion recognition, in figure \ref{fig:Concat}, results in degraded model performance. To address this issue, we propose a second stage that employs a Multi-modal Anchor Gated Transformer (MAGT) to effectively integrate features across the three modalities. Specifically, each modality serves as an anchor to aggregate complementary information from other modalities. Specifically, we first use the audio and video features as anchors to aggregate information from the other modalities. Given the strong performance of the text modality, the raw text features are used as an anchor to aggregate the audio and video features enriched by other modalities.

We construct the dataset at the conversation level, where utterance-level features from different modalities are arranged sequentially based on temporal order. For each utterance, speaker and positional embeddings are added. Speaker embedding uniquely maps each speaker in the dataset to a sequence ID, which is then embedded using an embedding layer.
\begin{equation}
    PE_{(pos,2i)}=sin(\frac{pos}{10000^{2i/d_{model}}}) \label{PE1}
\end{equation}
\begin{equation}
    PE_{(pos,2i+1)}=cos(\frac{pos}{10000^{2i/d_{model}}}) \label{PE2}
\end{equation}
\begin{equation}
    SE = Embedding[s_{u_1},s_{u_2}, \dots, s_{u_k}]
\end{equation}
\begin{equation}
    H_m = F_{m} + PE + SE
\end{equation}
where $PE$ is positional embedding and $SE$ is speaker embedding. $s_{u_k}$ represents speaker of utterance $u_k$. $F_m = (F_{m_1},F_{m_2}, \dots,F_{m_k})$ and $m \in \{t,a,v\}$. $H_m$ is the positional- and speaker-aware utterance sequence representation for $m$ modality.

Given the remarkable success of Transformers in NLP, the effectiveness of gating mechanisms for preserving salient features, and the superior performance of the text modality compared to audio and video in ERC tasks, we introduce an Anchor-Gated Transformer. This structure leverages a Transformer encoder with three inputs (Query $Q \in R^{l_{q} \times d_{k}}$, Key $K\in R^{l_{k} \times d_{k}}$, and Value $V\in R^{l_{k} \times d_{k}}$) to integrate features across modalities.
\begin{equation}
    H_{m \to m} = Transformer(H_m,H_m,H_m)
\end{equation}
\begin{equation}
    H_{n \to m} = Transformer(H_m,H_n,H_n)
\end{equation}
where $m \in \{t,a,v\}$ and $n \in \{t,a,v\}-\{m\}$. $H_{m \to m}$ represents anchor modality aggregating its own information. $H_{n \to m}$ represents an anchor modality aggregating information from other modalities.

To enhance the emotional representation of each modality, we incorporate a gating mechanism to filter out irrelevant information and retain the most effective emotional features. 
\begin{equation}
    \alpha_{n \to m} = \sigma(W_{n \to m} \cdot H_{n \to m} + b_{n \to m})
\end{equation}
\begin{equation}
    H_{n \to m}^{'} = H_{n \to m} \otimes \alpha_{n \to m}
\end{equation}
where $W_{n \to m}$ is a weight matrix, $b_{n \to m}$ is a bias parameter, $\alpha_{n \to m}$ represents gate, and $\otimes$ is the element-wise product.

\subsection{Emotion Classifier}
The emotion classification task is performed using a linear layer. The features $t'$, extracted from the Multi-modal Attention and Graph-based Transformer (MAGT), are transformed into the emotion label $p_i$ corresponding to each utterance $u_i$. 

The classification process can be formalized as follows:
\begin{equation}
p_i = \text{argmax}(\text{softmax}(W \cdot t' + b)),
\end{equation}
where $W \in \mathbb{R}^{C \times d}$ and $b \in \mathbb{R}^{C}$ are the weight matrix and bias vector of the linear layer, respectively. Here, $C$ denotes the number of emotion classes, and $d$ represents the dimension of the feature vector $t'$. The $\text{softmax}$ function ensures that the outputs are normalized probabilities across all emotion classes, and $\text{argmax}$ selects the class with the highest probability as the predicted label $p_i$.

\subsection{Training}

In the first stage, the utterance-level feature extraction is optimized using the loss function defined in Equation \ref{Loss1_all}. In the second stage, the multi-modal fusion is performed using the knowledge distillation (KD) loss functions defined in Equations \ref{loss_s} and \ref{loss_f}. These can be formalized as:
\begin{equation}
    L_{KD}^{i} = L_{f} + L_{s}
\end{equation}
where \(i\) refers to either the audio or video modality.

The total loss function in the second stage is the sum of the cross-entropy loss and the two distillation loss functions, weighted by their respective coefficients. This can be expressed as:
\begin{equation}
    L_{\text{total}} = L_{\text{CE}} + \alpha L_{\text{KD}}^\text{a} + \beta L_{\text{KD}}^\text{v}
\end{equation}
where \(\alpha\) and \(\beta\) are the coefficients for the distillation losses of the audio and video modalities, respectively.

\section{Experiments}
\subsection{Datasets}
In this section, we introduce two widely adopted benchmark datasets: MELD and IEMOCAP. Following descriptions for specific details of these two datasets:

\textbf{MELD} \cite{poria2018meld} is a multiparty conversation dataset containing over 1,400 dialogues and more than 13,000 utterances extracted from the TV show ``Friends.'' This dataset includes seven emotion categories: neutral, surprise, fear, sadness, joy, disgust, and anger.

\textbf{IEMOCAP} \cite{busso2008iemocap} consists of 7,433 utterances and 151 dialogues, divided into five sessions, each involving two speakers. Each utterance is labeled with one of six emotion categories: happiness, sadness, anger, excitement, frustration, and neutral. The training and development datasets are randomly split from the first four sessions in a 9:1 ratio. The test dataset comprises the last session.

\subsection{Experimental Setup}
\begin{table}[b]
\centering
\begin{tabular}{|c|c|c|c|}
\hline
\textbf{Stage}                 & \textbf{Parameter}             & \textbf{IEMOCAP} & \textbf{MELD} \\ \hline
\multirow{4}{*}{\makecell{Feature\\Extraction}}
                               & Dimensions                     & 768           & 768           \\ \cline{2-4} 
                               & Learning rate                  & 1e-5          & 1e-5          \\ \cline{2-4} 
                               & Batch                          & 4             & 4             \\ \cline{2-4} 
                               & Epochs                         & 10            & 10            \\ \hline
\multirow{4}{*}{\makecell{Multi-modal\\Fusion}}  
                               & Learning rate                  & 1e-5          & 1e-4          \\ \cline{2-4} 
                               & Batch                          & 16            & 16            \\ \cline{2-4} 
                               & Epochs                         & 30            & 30            \\ \cline{2-4} 
                               & $\alpha, \beta$                  & $0.7, 0.8$      & $0.01, 0.09$      \\ \hline
\end{tabular}
\caption{Hyperparameters used in the experiments.}
\label{tab:hyperparameters}
\end{table}

We evaluate the performance of our model on two datasets using Accuracy (Acc) and Weighted F1-score (W\_F1) as metrics. We designed a two-stage experimental process. The first stage focuses on extracting utterance-level features from different modalities, while the second stage involves multi-modal feature fusion using a dataset constructed at the conversation level. The hyperparameter settings are shown in Table \ref{tab:hyperparameters}. All experiments are conducted on a single NVIDIA GeForce RTX 2080 Ti GPU.

\subsection{Baselines}
We compare our proposed model, MAGTKD, against classic baselines, including DialogueRNN \cite{Poria_2019_dialoguernn}, DialogueGCN \cite{ghosal-etal-2019-dialoguegcn}, MMGCN \cite{hu-etal-2021-mmgcn}, and DialogueCRN \cite{hu-etal-2021-dialoguecrn}, as well as state-of-the-art models such as A-DMN \cite{xing-2022-A-DMN}, DialogueINAB \cite{kang-2023-Dialogueinab}, SACCMA \cite{GUO-2024-saccma}, Ada2I \cite{Nguyen-2024-Ada2I}, and GraphCPC \cite{Li-2024-GraphCFC}.

\subsection{Comparative Experiments}
\begin{table}[t]
    \centering
    \begin{tabular}{ccccc}
        \toprule
        \multirow{2}{*}{\textbf{Model}} & \multicolumn{2}{c}{\textbf{IEMOCAP}} & \multicolumn{2}{c}{\textbf{MELD}} \\
        \cmidrule(r){2-3} \cmidrule(r){4-5}
         & \textbf{Acc} & \textbf{W\_F1} & \textbf{Acc} & \textbf{W\_F1} \\
        \midrule
        DialogueRNN  & 63.4  & 62.75 & 60.31  & 57.66     \\
        DialogueGCN  & 65.25 & 64.18 & -     & 58.1  \\
        MMGCN        & 66.22 & -     & 58.65 & -     \\
        DialogueCRN  & 66.05 & 66.2  & 60.73 & 58.39 \\
        A-DMN        & 64.6  & 64.3  & -     & \underline{60.45} \\
        DialogueINAB  & 67.32 & 67.22 & 60.52 & 57.78 \\
        SACCMA    & 67.41 & 67.1  & 62.3  & 59.3  \\
        Ada2I  & 68.76 & \underline{68.97}  & \underline{63.03}  & 60.38  \\
        GraphCFC  & \underline{69.13} & 68.91  & 61.42  & 58.86  \\
        \midrule
        Ours        & \textbf{69.38} & \textbf{69.59} & \textbf{66.36} & \textbf{65.32} \\
        \bottomrule
    \end{tabular}
    \caption{Performance Comparison on IEMOCAP and MELD. Best results are in bold, second-best are underlined.}
    \label{tab:comparison}
\end{table}

Table~\ref{tab:comparison} compares our model with prior works on IEMOCAP and MELD. Our model achieves the best performance on both datasets, setting new state-of-the-art (SOTA) results. On IEMOCAP, we achieve 69.38\% accuracy and 69.59\% weighted F1 (W\_F1), outperforming GraphCFC by 0.99\% and 1.56\%, respectively. On MELD, our model achieves 66.36\% accuracy and 65.32\% W\_F1, with improvements of 5.17\% and 6.83\% over Ada2I, the previous SOTA. These results demonstrate our model's ability to effectively integrate multi-modal features and handle challenges in conversational emotion recognition through prompt learning, knowledge distillation, and advanced fusion techniques.

\subsection{Visualization and Analysis}
\begin{figure*}
    \centering
    \includegraphics[width=1\linewidth]{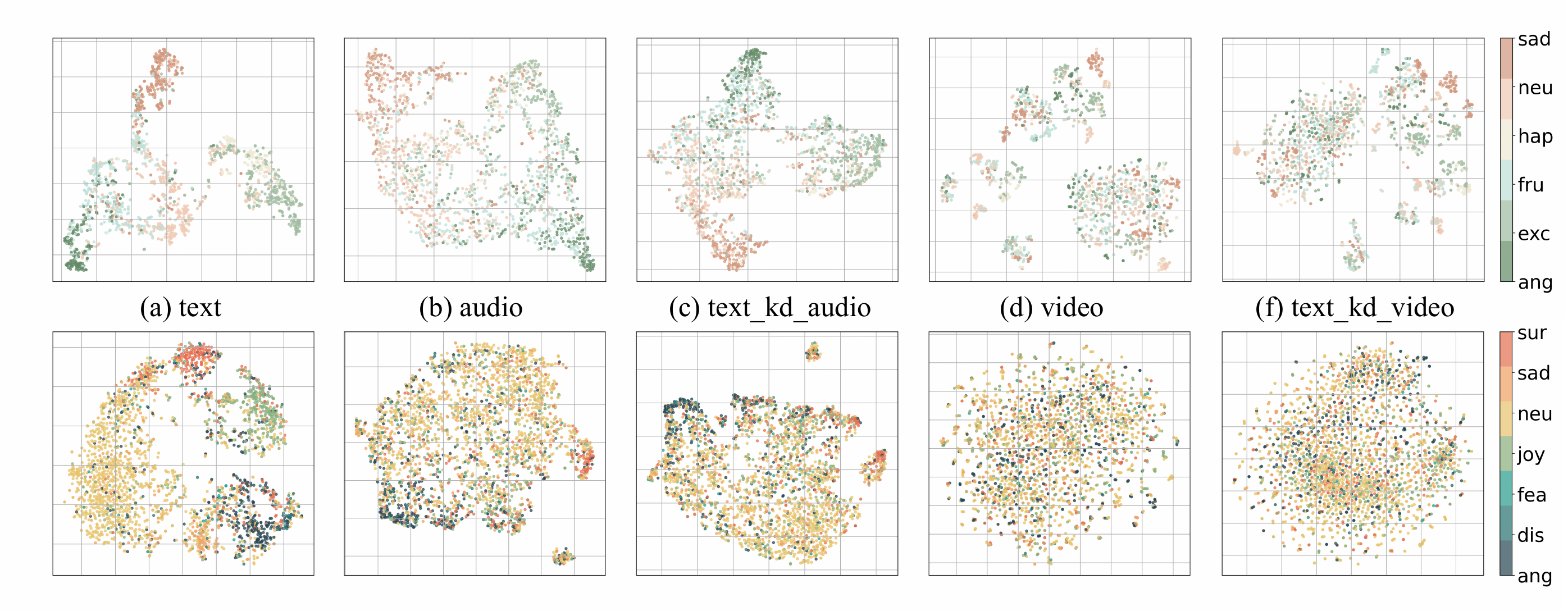}
    \caption{t-SNE visualization of feature representations for IEMOCAP (top row) and MELD (bottom row) datasets. ``kd'' refers to knowledge distillation.}
    \label{result_graph}
\end{figure*}
Figure~\ref{result_graph} shows t-SNE visualizations of the feature representations from the IEMOCAP and MELD datasets. We visualize single-modal features (text, audio, and video) as well as features enhanced by knowledge distillation, where the text modality guides the audio and video modalities.

The visualizations indicate that the text modality has the strongest discriminative power across both datasets, followed by audio, while video shows the least discriminative ability. After applying knowledge distillation, the audio modality improves significantly, benefiting from the text modality's guidance. However, the video modality struggles to learn effectively, highlighting the challenge of transferring knowledge to modalities with weaker feature representations.

Furthermore, the audio modality in IEMOCAP shows stronger discriminative power compared to MELD, which results in better performance when learning from the text modality. In the next sections, we will provide quantitative results to further explore this trend.

\subsection{Ablation Studies}

\begin{table}[t]
\centering
\begin{tabular}{ccccc}
\toprule
\multirow{2}{*}{\textbf{Module}} & \multicolumn{2}{c}{\textbf{IEMOCAP}} & \multicolumn{2}{c}{\textbf{MELD}} \\
\cmidrule(r){2-3} \cmidrule(r){4-5}
 & \textbf{Acc} & \textbf{W\_F1} & \textbf{Acc} & \textbf{W\_F1} \\
\midrule
\multicolumn{5}{c}{Feature Extraction} \\
\midrule
T      & 67.09 & 67.46 & 62.79 & 62.99 \\
A      & 47.01 & 45.91 & 50.38 & 44.80 \\
V      & 27.84 & 26.28 & 40.91 & 36.84 \\
$\text{A}_{\text{KD}}$ & 50.03 & 49.65 & 49.08 & 45.69 \\
$\text{V}_{\text{KD}}$ & 25.63 & 20.21 & 40.45 & 36.06 \\
\midrule
\multicolumn{5}{c}{Multi-modal Fusion} \\
\midrule
\multicolumn{5}{c}{Concat} \\
T+A+V            & 68.52 & 68.64 & 65.71 & 65.06 \\
T+A+$\text{V}_{\text{KD}}$       & 64.70 & 64.38 & 60.46 & 55.74 \\
T+$\text{A}_{\text{KD}}$+V        & 68.64 & 68.70 & 65.86 & 65.18 \\
T+$\text{A}_{\text{KD}}$+$\text{V}_{\text{KD}}$    & 65.37 & 65.19 & 60.61 & 55.67 \\
\midrule
\multicolumn{5}{c}{MAGT} \\
T+A+V            & 68.08 & 68.29 & 66.32 & 65.30 \\
T+A+$\text{V}_{\text{KD}}$       & 68.08 & 68.18 & 65.79 & 64.73 \\
T+$\text{A}_{\text{KD}}$+V        & \textbf{69.38} & \textbf{69.59} & \textbf{66.36} & \textbf{65.32} \\
T+$\text{A}_{\text{KD}}$+$\text{V}_{\text{KD}}$    & 68.88 & 69.02 & 65.79 & 64.73 \\
\bottomrule
\end{tabular}
\caption{Ablation studies on different modalities and fusion methods for IEMOCAP and MELD. ``KD'' indicates knowledge distillation, ``Concat'' is the simple fusion method, and ``MAGT'' is our proposed fusion method. Best results are in bold.}
\label{tab:module_comparison}
\end{table}

Table~\ref{tab:module_comparison} shows the results of ablation studies on feature extraction and multi-modal fusion. We evaluate individual modalities and the effects of knowledge distillation from the text modality to audio and video modalities.

The results highlight that knowledge distillation improves the performance of the audio modality but has little effect on the video modality. This aligns with the observations in Figure~\ref{result_graph}, where the video modality struggles to learn effectively from the text modality due to its weaker feature representation. On the other hand, the audio modality benefits more from the distillation process, particularly in the IEMOCAP dataset where it has a stronger feature representation compared to MELD.

For multi-modal fusion, we first tested the Concat method. The best performance was achieved by fusing the strongest modality features (text, distilled audio, and undistilled video). However, adding the distilled video modality degraded performance. We then tested the MAGT fusion method, which also performed best when fusing the strongest modalities. Notably, MAGT maintained stable performance even when the weakest modality, the distilled video, was included.

These results demonstrate that MAGT effectively integrates emotional cues from different modalities, even when some modalities contribute less useful information.

\subsection{Complexity Analysis}  
We define the feature dimensions of the prior frame-level model as \((S, L, D)\), and for the proposed utterance-level model as \((C, U, D)\), where:

\begin{itemize}  
    \item \(S\): number of samples,
    \item \(L\): frame sequence length,
    \item \(D\): hidden feature dimension,
    \item \(C\): number of conversations,
    \item \(U\): number of utterances per conversation.
\end{itemize}

Assuming \(C \cdot U = S\) and \(L\) is consistent across modalities.

\paragraph{Spatial Complexity}  
The frame-level model has spatial complexity:
\[
O(S \cdot L \cdot D) \quad \text{(linear with respect to \(S\) and \(L\))}.
\]
while the utterance-level model has:
\[
O(C \cdot U \cdot D) \quad \text{(linear with respect to \(C\) and \(U\))}.
\]
which reduces by a factor of \(L\) since \(C \cdot U = S\).

\paragraph{Temporal Complexity}  
The frame-level model’s temporal complexity is:
\[
O(S \cdot L^2 \cdot D) \quad \text{(quadratic with respect to \(L\))}.
\]
whereas the utterance-level model’s complexity is:
\[
O(C \cdot U^2 \cdot D) \quad \text{(quadratic with respect to \(U\))}.
\]
Thus, the relative temporal complexity is:
\[
\frac{O(C \cdot U^2 \cdot D)}{O(S \cdot L^2 \cdot D)} = \frac{U}{L^2}.
\]
For shorter dialogues (\(U \ll L\)), the proposed model has a significant reduction in complexity. Additionally, spatial complexity is reduced by a factor of \(L\). In summary, the proposed model is more efficient, with linear spatial and quadratic temporal complexity in \(U\), compared to the frame-level model’s quadratic temporal complexity in \(L\).

\subsection{Hyper-parametric Analysis}
\begin{figure}
    \centering
    \includegraphics[width=1\linewidth]{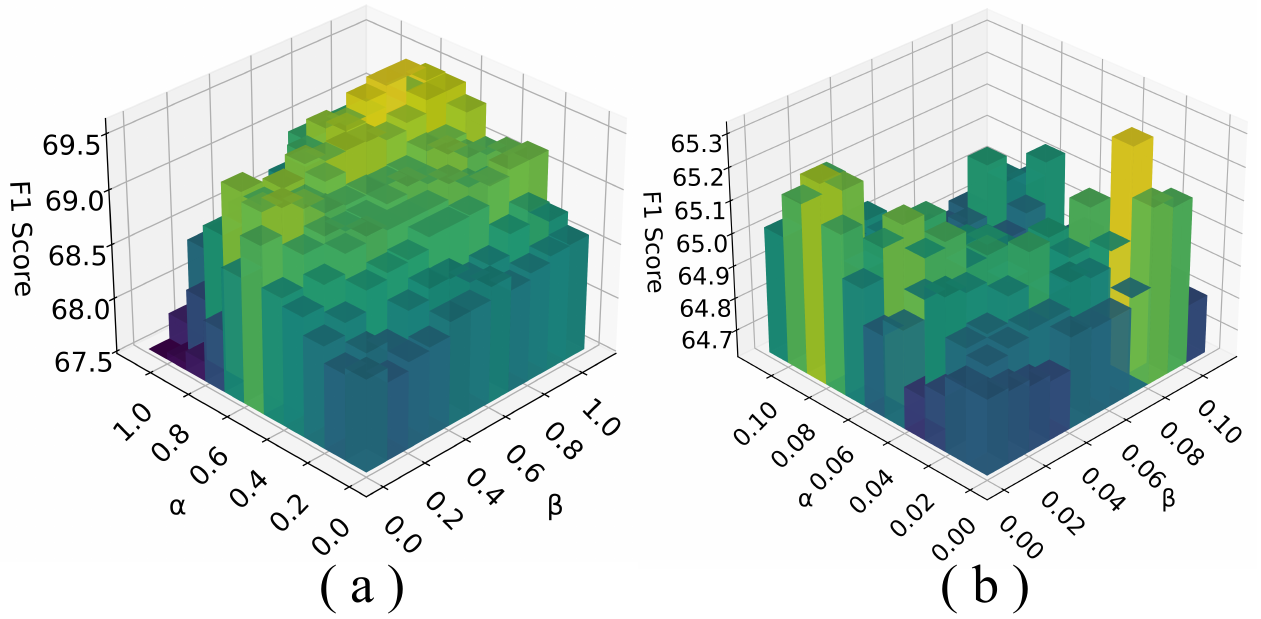}
    \caption{(a) Hyperparametric analysis for IEMOCAP, (b) Hyperparametric analysis for MELD. Here, \(\alpha\) and \(\beta\) are the coefficients for audio and video knowledge distillation losses, respectively.}
    \label{Hyper-parametric Analysis}
\end{figure}
Figure~\ref{Hyper-parametric Analysis} shows the effect of varying the hyperparameters \(\alpha\) (audio distillation coefficient) and \(\beta\) (video distillation coefficient) on model performance for IEMOCAP and MELD. For IEMOCAP, changing \(\alpha\) significantly affects performance when \(\beta\) is fixed, while adjusting \(\beta\) with a fixed \(\alpha\) results in smaller variations. This suggests that the audio modality better benefits from knowledge distillation, whereas the video modality shows weaker learning. This aligns with the observations in Figure~\ref{result_graph}, where the audio modality has better feature discriminability than the video modality. In MELD, setting \(\alpha = 0\) and increasing \(\beta\) initially improves performance, but further increases lead to a decline. Similarly, setting \(\beta = 0\) and varying \(\alpha\) shows an initial performance boost followed by a decrease, confirming that the KD loss improves model performance but requires careful tuning.

\subsection{Convergence Analysis}
\begin{figure}
    \centering
    \includegraphics[width=1\linewidth]{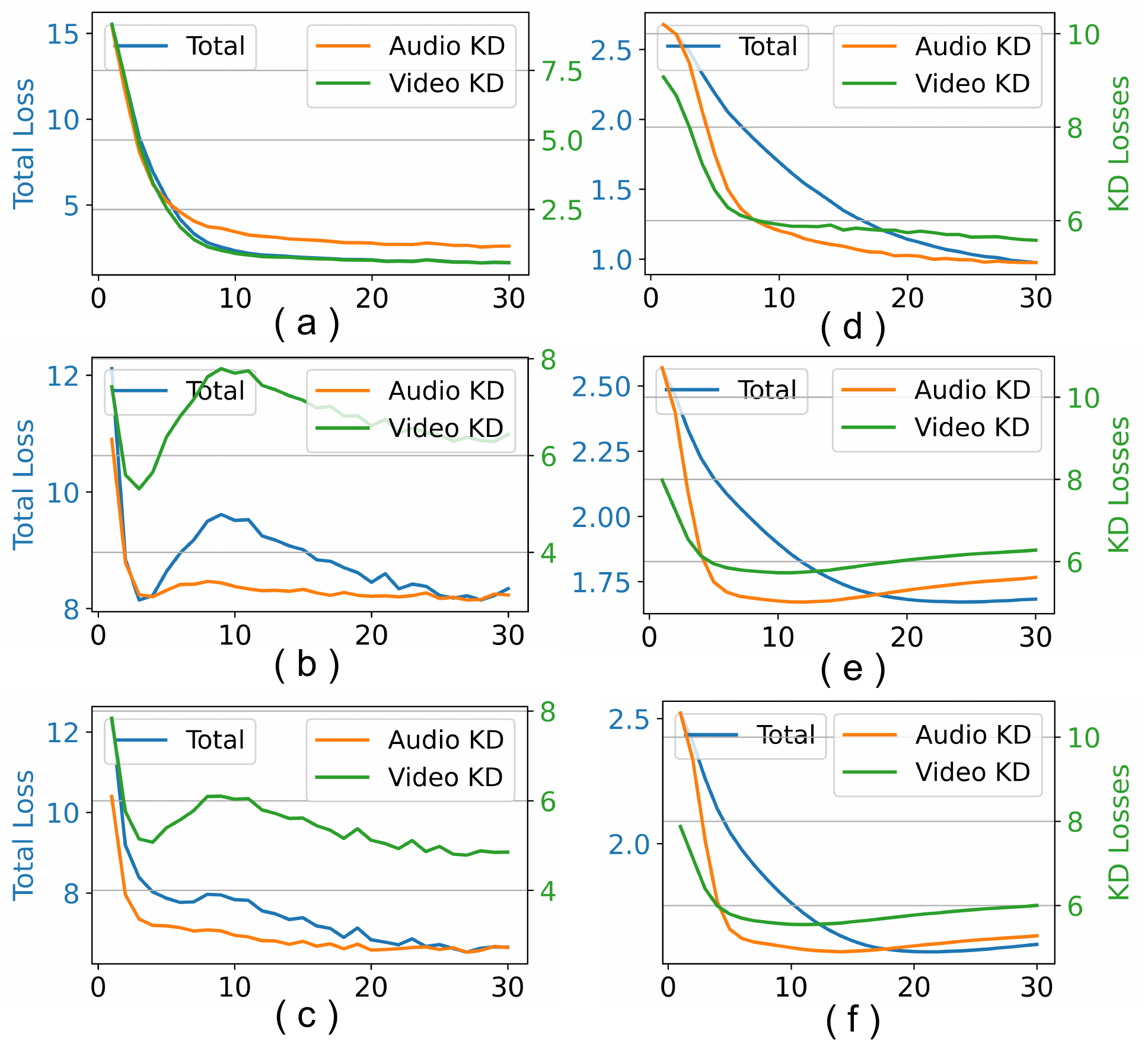}
    \caption{(a)-(b) show the variations in Total Loss and Knowledge Distillation (KD) Loss over training epochs for IEMOCAP across training, validation, and test sets. (d)-(f) show the same for MELD.}
    \label{Convergence Analysis}
\end{figure}
Figure~\ref{Convergence Analysis} presents the convergence behavior of the model on IEMOCAP and MELD datasets. For IEMOCAP (a)-(c), as the number of epochs increases, both the total loss and the KD loss for audio and video modalities converge. The KD loss for the audio modality converges steadily, while the video modality experiences fluctuations before stabilizing, which is consistent with the lower discriminability of video features, as shown in Figure~\ref{result_graph}. Similar patterns are observed for the validation and test sets. For MELD (d)-(f), the trends are similar, with both the total loss and the KD losses for audio and video modalities converging as epochs increase. These results confirm the effectiveness of our model and the positive impact of knowledge distillation on training stability and performance.

\section{Conclusion}

The proposed MAGTKD model effectively addresses the challenges of multi-modal ERC by leveraging prompt learning to extract robust textual representations and employing knowledge distillation to enhance weaker modalities. The subsequent use of MAGT enables efficient aggregation of emotional information across modalities, resulting in state-of-the-art performance on both the MELD and IEMOCAP datasets. Future work will explore extending MAGTKD to handle more complex multi-modal scenarios, such as incorporating dynamic contextual information in real-time conversations or addressing challenges posed by highly imbalanced datasets. 

\section*{Acknowledgements}
This work was supported by the National Natural Science Foundation of China (Grant no.62276265 and 62406326).

\bibliographystyle{named}
\bibliography{reference}

\end{document}